\documentclass[11pt]{article}
\usepackage{nodalida2025}
\usepackage{times}
\usepackage{url}
\usepackage{latexsym}

\usepackage{xspace}
\usepackage{booktabs}
\usepackage{multirow, makecell}
\usepackage{graphicx}
\usepackage{covington}
\usepackage{relsize}
\usepackage{comment}
\usepackage{placeins}
\usepackage{float}
\usepackage{xcolor}
\definecolor{darkblue}{rgb}{0, 0, 0.5}
\usepackage{hyperref}
\hypersetup{
           breaklinks=true,   
           colorlinks=true,   
           pdfusetitle=true,  
           citecolor=darkblue, 
           linkcolor=darkblue, 
           urlcolor=darkblue
        }

\newcommand{\norpac}{{NorPaC}\xspace}
\newcommand{\norec}{{NoReC}\xspace}
\newcommand{\norecfine}{{NoReC\textsubscript{\textit{fine}}}\xspace}
\newcommand{\norecsentence}{{NoReC\textsubscript{\textit{sentence}}}\xspace}
\newcommand{\fen}{F\textsubscript{1}\xspace}

\newcommand{\bestnorpac}[1]{\textbf{#1}}
\newcommand{\bestboth}[1]{\textbf{#1}}

\aclfinalcopy 

\title{Mixed Feelings:\\ Cross-Domain Sentiment Classification of Patient Feedback}

\author{Egil Rønningstad\thanks{~~The authors contributed equally.} , Lilja
  Charlotte Storset\footnotemark[1] , Petter Mæhlum, Lilja Øvrelid, Erik
  Velldal \\ Department of Informatics, University of Oslo, Norway\\
  {\tt \{egilron, liljacs, pettemae, liljao, erikve\}@uio.no}}

\date{}

\begin{document}
\maketitle

\begin{abstract}
Sentiment analysis of patient feedback from the public health domain can aid decision makers in evaluating the provided services. 
The current paper focuses on free-text comments in patient surveys about general practitioners and psychiatric healthcare, annotated with four sentence-level polarity classes -- positive, negative, mixed and neutral -- while also attempting to alleviate data scarcity by leveraging general-domain sources in the form of reviews. 
For several different architectures, we compare in-domain and out-of-domain effects, as well as the effects of training joint multi-domain models.
\end{abstract}

\section{Introduction}
\label{sec:intro}
Sentiment analysis (SA), the computational analysis of opinions and emotions expressed in text, is one of the applications of natural language processing (NLP) that have found the most wide-spread use across many different areas, including medical domains \citep{yadav-etal-2018-medical}. As the task is mostly approached as one of supervised learning, access to sufficient amounts of labeled data is the main driver of performance. However, as manual annotation is costly, 
labeled data also  represents a main bottleneck. For this reason it is typically desirable to be able to reuse existing resources when developing SA tools for a new area of application. Unfortunately, domain-sensitivity is a well-known effect across many different NLP tasks. Models trained on data from one domain (or genre or text-type) often underperform when applied to another due to variations in language use, terminology, and contextual nuances \citep{7891035, GräßerFelix2018ASAo}.

This paper investigates cross-domain effects in polarity classification of public health data, more specifically free-text comments from patient surveys for general practitioners and psychiatric healthcare providers. We here investigate the usefulness of  data from a different domain and genre, i.e. professionally authored reviews collected from Norwegian news publishers. The datasets are annotated at the sentence level with the same four-class polarity labels; positive, negative, mixed, and neutral. In the following, we compare non-neural and neural architectures in both in-domain and cross-domain settings with the goal of providing high-quality sentiment analysis for Norwegian patient comments.



\section{Datasets}
\label{sec:data}
We here briefly describe the two annotated SA datasets that form the basis of our experiments, also discussing some of their key differences. 

\paragraph{\norpac} 
For the health domain we will be using a dataset introduced by \citet{maehlum-etal-2024-difficult}, comprising free-text comments from surveys conducted by the Norwegian Institute of Public Health (NIPH), as part of their so-called patient-reported experience measures (PREMs). The dataset is dubbed \norpac\ -- short for Norwegian Patient Comment corpus -- and comprises two related sub-domains, corresponding to feedback on General Practitioners (GPs) and Special Mental Healthcare (SMH), with a total of 7693 sentences (4002 from GP and 3691 from SMH) annotated for polarity. 

The \norpac dataset is a valuable accession to Norwegian corpora, as it gives valuable insights to the national public health system. The texts are written by patients after encounters with the healthcare system, and gives rise to language with an everyday character, such as sentences with a conversational tone or even incomplete sentences and spelling mistakes. Example \ref{ex:niph_typo_caps} shows a positive patient feedback sentence that is written solely in capital letters, in addition to containing a typing mistake in the personal pronoun \textit{jeg}, 'I'. Example \ref{ex:niph_too_bad} shows a negative review with a colloquial tone, containing three exclamation marks at the end of the utterance.

\begin{covexample}
\gll FIKK HENVISNING DA JGE BA OM DET, OG GÅR STADIG TIL UTREDNING DER.
Got referral when (I) asked about it, and goes constant to examination there.
\glt‘Got a referral when I asked for it, and am constantly going for examination there.’
\glend
\label{ex:niph_typo_caps}
\end{covexample}

\begin{covexample}
\gll Det er for dårlig!!!
It is too bad!!!
\glt‘It is too bad!!!’
\glend
\label{ex:niph_too_bad}
\end{covexample}

\paragraph{\norec}
The Norwegian Review Corpus \citep[\norec;][]{velldal-etal-2018-norec} comprises full-text reviews collected from major Norwegian news sources, covering a range of different domains (movies, music, literature, restaurants, various consumer products, etc.). We here  use a version dubbed NoReC$_{fine}$ \cite{ovrelid-etal-2020-fine}, a subset of roughly 11,000 sentences across more than 400 reviews with fine-grained sentiment annotations, here aggregated to the sentence-level \cite{kutuzov-etal-2021-large} using the above-mentioned label set of four classes.\footnote{\url{https://huggingface.co/datasets/ltg/norec_sentence}} In contrast to \norpac, the reviews are written by professional authors, meaning more creative writing but with sentences that are typically complete and grammatically correct. 

\begin{covexample}
\gll Den er en pølse i salatens tid, en slags mumlemanisk manns-modernitets-manifestasjon
It is a sausage in salad's.the time, a kind.of mumblemaninc man-modernity-manifestation
\glt‘It is a sausage in the age of the salad, a kind of mumble-manic male-modernity-manifestation’
\glend
\label{ex:norec_1}
\end{covexample}

Example \ref{ex:norec_1} shows one of many creative sentences in \norec. \textit{En pølse i salatens tid}, `a sausage in the age of the salad', is a figurative way to emphasize the fact that this movie is not among the trendy, i.e. `the salad', but rather acts like `a sausage'. Further, the author describes the movie as a \textit{mumlemanisk manns-modernitets-manifestasjon}, 'mumble-manic male-modernity-manifestation'. This exemplifies the complexity of many of the texts in the NoReC dataset where authors may construct new and creative expressions.

\paragraph{Genre and text type}
The two datasets can be said to be found at opposite ends in terms of language and writing style. In contrast to the professionally authored reviews in \norec, containing grammatically correct texts with higher complexity and creativity, the \norpac patient comments consist of more colloquial language. It also comes with many of the other hallmarks of user-generated content, such as more frequent spelling mistakes and incomplete sentences, as well as unorthodox use of case and punctuation. While such properties will generally contribute to increasing the vocabulary size, \norec still contains almost three times as many unique lemmas as \norpac, due to the fact that it contains more creative and varied language (with a higher degree of figurative expressions, etc.), as mentioned above, in addition to covering multiple domains.

\begin{table*}[ht!]
    \centering
    \smaller
    \begin{tabular}{@{}clrrrrr@{}}
    \toprule 
              && Positive & Negative & Neutral & Mixed & Total\\
      \cmidrule(l){3-7}
    \multirow{3}{*}{\textbf{GP}} 
    & Sentences & 1265 (32\%) & 1903 (48\%) & 654 (16\%) & 174 (4\%) & 4002\\
    & Avg. tokens & 11.8 & 15.61 & 10.38 & 19.99 & 13.81\\
      \cmidrule(l){3-7}
    \multirow{3}{*}{\textbf{SMH}}
    & Sentences & 1524 (41\%) & 1604 (44\%) & 291 \hphantom{0}(8\%)& 266 (7\%)& 3691\\
    & Avg. tokens & 13.1 & 18.48 & 10.53 & 23.68 & 15.94\\
      \cmidrule(l){3-7}
    \multirowcell{3}{\textbf{\norpac}\\ (GP+SMH)}  
    & Sentences & 2789 (36\%) & 3507 (46\%) & 945 (12\%) & 440 (6\%) & 7693\\
        & Avg. tokens & 12.53 & 17.03 & 10.78 & 22.29 & 14.93\\
      \cmidrule(l){3-7}
    \multirow{3}{*}{\textbf{NoReC}} 
    & Sentences & 3514 (31\%) & 1663 (15\%) & 5393 (47\%) & 867 (8\%) & 11437\\
    & Avg. tokens & 18.57 & 18.18 & 13.78 & 25.92 & 16.78 \\
    \bottomrule 
    \end{tabular}
    \caption{For each polarity class we show the distribution of number of sentences and average sentence length across the GP and SMH datasets within \norpac, and for the \norec dataset.}
    \label{tab:datasets_stats}
\end{table*}

\paragraph{Class distribution}  
Table~\ref{tab:datasets_stats} summarizes some relevant statistics for the two corpora, showing the number of examples across the four classes, as well as average token length of sentences. 

For the \norec reviews, we see that we have many more examples for the positive than the negative category. For the \norpac patient feedback, in contrast, the negative category is notably larger, although the number of positive and negative examples are more balanced than in \norec. 

Another striking difference is the much higher ratio of neutral sentences in \norec compared to \norpac; 47\% vs. 12\%, respectively. This is not surprising if we consider the genre differences; professional reviews need to provide a lot of non-sentiment bearing background and descriptions of the object under review. 
The ratio of sentences with mixed polarity, however, is similar across the datasets, and is also the smallest sentiment class.

Related to the class distribution, we also observe some interesting differences with respect to the average token length of sentences. While the length is the same across the positive and negative sentences in the \norec reviews, the length of negative sentences in the \norpac patient comments tend to be substantially longer than the positive ones. However, for both datasets we see that neutral sentences tend to be shorter, while the mixed class displays substantially longer average length, which intuitively makes sense given that they per definition must express at least two opposing sentiments.

\section{Experimental results}
\label{sec:experiments}

Below we report experimental results for a range of different models and architectures on the datasets described above. We start by providing 
details about the models and the experimental set-up, before 
discussing the results for both in-domain and cross-domain classification.

\subsection{Models and experimental set-up}
\label{sec:models}

The NorBERT3 series of models \cite{samuel-etal-2023-norbench,kutuzov-etal-2021-large} represent the 3rd generation of pre-trained Norwegian masked language models (MLMs) based on the BERT transformer architecture \cite{devlin-etal-2019-bert}. We fine-tuned text classifiers for two different sizes of NorBERT3 -- Base and Large -- with 123M and 353M parameters, respectively. GPU memory requirements were 8 and 35 GB. 
The NorT5 \cite{samuel-etal-2023-norbench} models are pretrained on the same Norwegian data as NorBERT3, and we fine-tune NorT5 Large to generate sentiment labels as a sequence-to-sequence task. NorT5 Large has 808M parameters. During fine-tuning with a batch size of 24, 71GB GPU memory was used.
For all these models we report the mean weighted average \fen over 3 runs. More details of the hyperparameter search are found in Appendix \ref{sec:appendix-hptuning}.
As a baseline, we also train a Support Vector Machine (SVM) model with a linear kernel and bag-of-words features.\footnote{The features correspond to the full vocabulary of the tokenized texts for each corpora, as preliminary experiments showed that best results were obtained without any feature selection or weighting (i.e. no TF-IDF, frequency cutoffs, etc.). The number of features range from approximately 5K for the GP/SMH models, through 8K for the full \norpac data and 22K for \norec, and finally 27K for \norpac\!+\norec.}  
The random baseline for the task yields an F1-score of between 22\% and 23\% for all training datasets, averaged across 1000 runs.

For \norec we use the predefined data split, with 80-10-10 percentages respectively for the training, validation and test set. We define a similar split for \norpac, randomly selected on the comment-level to make sure sentences from the same comment are not separated across splits, while also ensuring a balanced class distribution.



\subsection{In-domain patient comment results}

Table~\ref{tab:norpac-results} shows results when training and testing on sentences from the \norpac corpus. While the main focus of this section is to assess the in-domain performance of models trained on the \norpac patient comments, recall that this corpus comprises two different sources; feedback regarding General Practitioners (GPs) and Special Mental Healthcare (SMH). We therefore also report results for training and testing on data from the individual sources separately -- including cross-source training and testing. 
\begin{table}[tb] 
\smaller
    \centering
    \begin{tabular}{@{}llccc@{}}
    \toprule 
    & & \multicolumn{3}{c}{Test}\\
    \cmidrule(l){3-5} 
    Model & Train & GP & SMH & \norpac \\
    \midrule
    \multirowcell{3}{SVM\\(BoW)}
    & GP  &\bestnorpac{63.65} &66.42 &64.96 \\
    & SMH &57.86 &66.77 &62.26 \\
    & \norpac &62.90&\bestnorpac{68.34}&\bestnorpac{65.52}\\
    \midrule
    \multirowcell{3}{NorBERT3\\(Base)}
 & GP   & \bestnorpac{84.13} &           82.02            & 83.14 \\
 & SMH  &           79.43  &           82.96            & 81.22 \\
 & \norpac &        83.61  & \bestnorpac{83.34} & \bestnorpac{83.49} \\
    \midrule
    \multirowcell{3}{NorBERT3\\(Large)}
 & GP   &           85.79  & \bestnorpac{84.85} & \bestnorpac{85.41} \\
 & SMH  &           81.23  &           84.61  &           82.95 \\
 & \norpac & \bestnorpac{86.00} &           84.28  &           85.22 \\
     \midrule
    \multirowcell{3}{NorT5\\(Large)}
 & GP & 84.34 & 83.65 & 84.08 \\
 & SMH & 81.03 & 84.24 & 82.70 \\
 & \norpac & \bestnorpac{85.03} & \bestnorpac{85.05} & \bestnorpac{84.54} \\
    \bottomrule
    \end{tabular}
    \caption{Results for training and testing on the GP and SMH datasets within NorPaC. 
    } 
    \label{tab:norpac-results}
\end{table} 

We see that training on GP yields very strong test results: Not only are in-domain results for training and testing for SMH lower, but test results on SMH are competitive when training on GP. 
In the same vein, we see that for most models, joint training on the entire \norpac data boosts results for SMH, with the only exception being NorBERT3 Large, where the best results for SMH are actually found when training on GP only (although the differences are marginal). In sum, we find that, within the \norpac domain(s), the generalization capabilities of the GP-trained models are so good that the benefit of joint training on GP and SMH are less than anticipated. One contributing factor here might be that the GP data overall is written in a more explicit and straightforward manner compared to the SMH data, which might contain parts that are perceived as noisy for the model. Hence, training on GP and testing on PHV yields better results than vice versa. Finally, and as expected, we see that the neural models outperform the SVM model and that larger models generally tend to outperform smaller ones, although NorT5 Large actually tends to be outperformed by the smaller NorBERT3 Large model.



\subsection{Cross-domain results}

\begin{table}[tb!] 
    \centering
    \smaller 
    \begin{tabular}{@{}llcc@{}}
    \toprule 
    & & \multicolumn{2}{c}{Test}\\
    \cmidrule(l){3-4} 
    Model & Train & \norpac & \norec\\ 
    \midrule
    \multirowcell{3}{SVM\\(BoW)}
    & \norpac       &        65.52 &         37.84\\ 
    & \norec      &          42.11 &         \textbf{54.42}\\ 
    & \norpac\!+\norec &\bestboth{66.20}&    53.35\\
    \midrule
    \multirowcell{3}{NorBERT3\\(Base)}
 & \norpac       &           83.49 &           59.09 \\
 & \norec      &           68.03 &           75.63 \\ 
 & \norpac\!+\norec & \bestboth{83.71}& \bestboth{76.14} \\ 
    \midrule
    \multirowcell{3}{NorBERT3\\(Large)}
& \norpac       & \bestboth{85.22}&           59.19 \\ 
& \norec      &           66.38 & \bestboth{78.88} \\ 
& \norpac\!+\norec &           85.03 &           78.40 \\ 
    \midrule
    \multirowcell{3}{NorT5\\(Large)}
& \norpac  & 84.54 & 58.14 \\
& \norec  & 70.88 & \bestboth{76.73} \\
& \norpac\!+\norec & \bestboth{85.06} & 75.79 \\

    \bottomrule
    \end{tabular}
    \caption{Results for training and testing on sentences from both the \norpac patient comments and the \norec reviews. }
    \label{tab:cross-results}
\end{table}

Table \ref{tab:cross-results} shows results for several combinations of training and testing on both \norpac and \norec. First, we note that the in-domain results for \norpac are substantially higher than the in-domain results for \norec. This makes sense, given that \norec in practice covers many different domains and has a much more diverse vocabulary than \norpac. This observation most likely also has bearings on the cross-domain results, where we see a smaller relative drop in performance when testing the \norec-trained models on \norpac, than vice versa.  
Another contributing factor to the (expected)  drops in performance for the cross-domain results can be the differences in the class distribution for the two datasets, as discussed above. 

Turning to the joint training on the combination of \norec and \norpac, we again see that the test scores are substantially higher on \norpac than \norec for all models. For the NorBERT3 Base model, the joint training improves results across both datasets. However, for NorBERT3 Large, we see that the in-domain variants gives the highest scores for both datasets, but only by a small margin. For the SVM model, we see the same tendency with in-domain training on \norec, yielding slightly better performance than joint training.

In an error analysis of in-domain vs. out-of-domain results for NorBERT3 Large evaluated on the \norpac test set, we observe that the model trained on \norpac is better at predicting negative sentences, compared to the model trained on \norec. Here, the in-domain model classifies 92\% of the negative samples correctly, whereas the out-of-domain model only identifies 39\% of them. Out of the true negative samples, the \norec-trained model predicts 59\% of them as neutral. We believe the prediction of the negative class is the largest contributor to the lower performance of the \norec-trained model, as this class makes up 46\% of the \norpac test set. However, there is one class for which this model performs slightly better than the in-domain model. As we recall from Table \ref{tab:datasets_stats}, the neutral class is the largest class in the \norec dataset. This is most likely the reason why the \norec-trained model classifies 95\% of these instances in the \norpac test set correctly, as opposed to the \norpac trained model, which correctly classifies 69\% of them. 
In sum, a closer look at the per-class results reveals clear effects of the class distribution in the training set on model performance.

\begin{figure}[bt]

    \centering
        \includegraphics[width=1\linewidth]{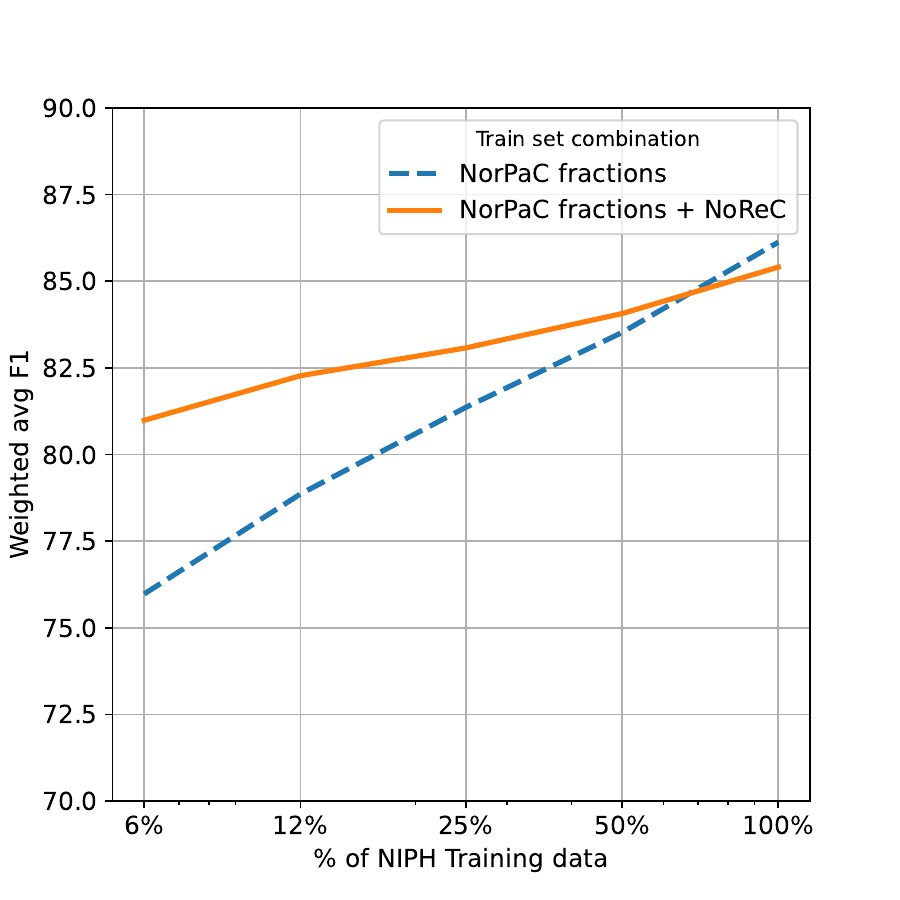}
    \caption{
    Learning curves, for two configurations: 
    \textbf{\norpac fractions}: The model is trained on fractions of the \norpac training split, from 6.25\% $\approx$ 6\% (386 samples) successively doubling the training set up to the full \norpac training split. \textbf{\norpac fractions + \norec}: The same fractions of the \norpac training split, mixed with the full \norec training split. All evaluations are on the full \norpac test set, averaged over three runs with different seeds, and with the amounts of in-domain training data shown on log-scale.}
    \label{fig:learning-curves}
\end{figure}

\paragraph{Learning curves for in-domain data} 
To gauge the effect of the number of in-domain training examples, we computed learning curves where models are trained on partitions that are created by successively halving the \norpac training set, 
with and without including the full NoReC training data. 
Figure~\ref{fig:learning-curves} plots the effect on finetuning NorBERT3 Large. 
Utilizing only 6.25\% (386 samples) of the \norpac training set we find a strong performance gain of adding the cross-domain \norec dataset. The effect is reduced, but present up to 50\% (3087 samples). However, with the full \norpac training set containing 6175 samples, we find that adding cross-domain data is harmful for the model performance. This shows how cross-domain data can help when in-domain datasets are small, but should not be added indiscriminately. 

\section{Summary 
}
\label{sec:summary}
This paper has reported experimental results for polarity classification of sentences in a Norwegian dataset dubbed \norpac, comprising free-text comments from patient surveys collected as part of evaluating public healthcare services. In addition to assessing cross-domain effects between two healthcare sub-domains -- feedback on general practitioners and psychiatric healthcare -- we have also assessed the effect of leveraging general-domain sentiment annotations, based on the NoReC review data. Rather than just annotating the simple binary classification of positive/negative sentences, our datasets additionally indicate both neutral and mixed sentences. We show how several of our tested model configurations surpass 85\% weighted \fen for this four-class set-up. We also show how including out-of-domain data improves model performance when in-domain data is limited, but that better performance can be achieved with in-domain data alone once the the amount of annotated data crosses a critical threshold. Our analyses give new insights into both the \norpac and \norec datasets, including the differences and similarities between them.

\section*{Acknowledgments}
This work was supported by two research projects funded by the Research Council of Norway (RCN), namely `Sentiment Analysis for Norwegian Text' (SANT), funded by an IKTPLUSS grant from  RCN (project no. 270908), and `Strengthening the
patient voice in health service evaluation: Machine learning on free-text comments from surveys and online sources', funded by a HELSEVEL grant from RCN (project no. 331770). Moreover, the computations were performed on resources provided through Sigma2 -- the national research infrastructure provider for High-Performance Computing and large-scale data storage in Norway.

\bibliographystyle{acl_natbib}
\bibliography{anthology_0, anthology_1, nodalida2025}

\clearpage 
\appendix
\section{Hyperparameter tuning for NorBERT3-based models}
\label{sec:appendix-hptuning}

We chose NorBERT3 base and large as the models to fine-tune for the text classification task. This model series has proven to perform well on previous comparisons for sentiment analysis on Norwegian sentences \citep{samuel-etal-2023-norbench}.
In order to find the best hyperparameters for our task, we first experimentally determine the best combination of learning rate and batch size. Table~\ref{tab:lr_bs} shows the results for the two model sizes. All experiments are evaluated by  accuracy on the development split, using the best of 10 epochs and one seed per hyperparameter combination. With the best performing settings for learning rate and batch size, we further search for improved performance by adjusting dropout in the classifier head, warm-up ratio and weight decay during fine-tuning. The search space for these hyperparameters are shown in Table~\ref{tab:searchspace}. The best performing settings are shown in Table~\ref{tab:dropout_w_w}. The final choice of hyperparameters are shown in Table~\ref{tab:BERT_hyperparameters}.

\begin{table}[h!]
    \centering
    \smaller 
\begin{tabular}{@{}lrrrr@{}} 
\toprule
Model & lr & 16 & 32 & 64 \\
\midrule
base & 1e-05 & \textbf{78.28} & 77.82 & 77.76 \\
base & 2e-05 & 77.92 & 78.12 & 77.79 \\
base & 5e-05 & 76.09 & 77.04 & 78.18 \\
\midrule
large & 1e-05 & 80.44 & 81.12 & 80.37 \\
large & 2e-05 & 80.96 & \textbf{81.35} & 80.89 \\
large & 5e-05 & 79.23 & 80.44 & 80.60 \\
\bottomrule
\end{tabular}
    \caption{Learning rate and batch size hyperparameter search for NorBERT3-base and large.}
    \label{tab:lr_bs}
\end{table}

\begin{table}[h!]
    \centering
    \smaller 
\begin{tabular}{@{}lr@{}}
\toprule
Model & Search space \\
\midrule
classifier dropout & [0.05, 0.1, 0.25, 0.4] \\
warm-up ratio & [0.01, 0.05, 0.1, 0.2] \\
weight decay & [0.001, 0.01, 0.1] \\
\bottomrule
\end{tabular}
    \caption{ Search space for classifier dropout, warmup ratio and weight decay for NorBERT3 base and large, after best learning rate and batch size was identified.}
    \label{tab:searchspace}
\end{table}

\begin{table}[H] 
    \centering
    \smaller 
    \begin{tabular}{@{}lrrrr@{}}
    \toprule
    Model & Dropout & Wu\_ratio & W\_decay & Dev acc. \\
    \midrule
    base & 0.25 & 0.20 & 0.010 & \textbf{78.77\%} \\
    base & 0.10 & 0.20 & 0.010 & 78.71\% \\
    base & 0.25 & 0.20 & 0.100 & 78.58\% \\
    \midrule
    large & 0.25 & 0.10 & 0.100 & \textbf{82.10\%} \\
    large & 0.25 & 0.10 & 0.001 & 81.91\% \\
    large & 0.40 & 0.20 & 0.001 & 81.84\% \\
    \bottomrule
    \end{tabular}
    \caption{Top-3 performing models, for NorBERT3 base and large, when searching for optimal parameters for classifier dropout, warm-up ratio and weight decay.}
    \label{tab:dropout_w_w}
\end{table}

\begin{table}[H] 
    \centering
    \smaller 

    \begin{tabular}{@{}lll@{}}
    \toprule
    Model & Base & Large \\
    \midrule
    batch size & 16 & 32 \\
    learning rate & 1e-05 & 2e-05 \\
    classifier dropout & 0.25 & 0.25 \\
    warmup ratio & 0.20 & 0.10 \\
    weight decay & 0.01 & 0.10 \\
    \bottomrule
    \end{tabular}
    \caption{Final hyperparameters selected for the NorBERT3 base and large finetuning, as informed by our hyperparameter search. Other hyperparameters are left as their defaults.}
    \label{tab:BERT_hyperparameters}
\end{table}

\end{document}